# Extreme 3D Face Reconstruction: Seeing Through Occlusions

Anh Tuấn Trần[1], Tal Hassner[2,3], Iacopo Masi[1], Eran Paz[3], Yuval Nirkin[3], and Gérard Medioni[1]
[1] Institute for Robotics and Intelligent Systems, USC, CA, USA
[2] Information Sciences Institute, USC, CA, USA
[3] The Open University of Israel, Israel

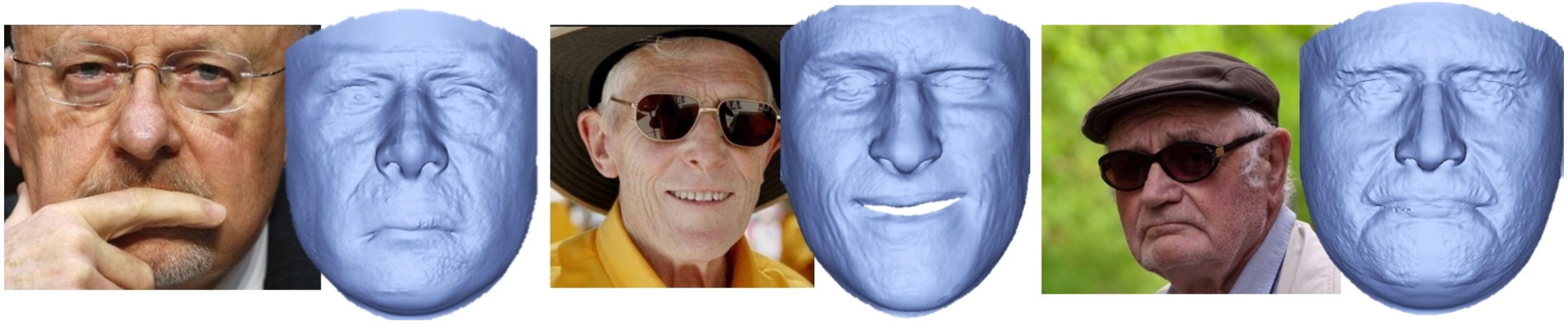

**Figure 1: Results of our method.** Detailed, complete 3D reconstructions shown next to their partially occluded input faces.

## Abstract

*Existing single view, 3D face reconstruction methods can produce beautifully detailed 3D results, but typically only for near frontal, unobstructed viewpoints. We describe a system designed to provide detailed 3D reconstructions of faces viewed under extreme conditions, out of plane rotations, and occlusions. Motivated by the concept of bump mapping, we propose a layered approach which decouples estimation of a global shape from its mid-level details (e.g., wrinkles). We estimate a coarse 3D face shape which acts as a foundation and then separately layer this foundation with details represented by a bump map. We show how a deep convolutional encoder-decoder can be used to estimate such bump maps. We further show how this approach naturally extends to generate plausible details for occluded facial regions. We test our approach and its components extensively, quantitatively demonstrating the invariance of our estimated facial details. We further provide numerous qualitative examples showing that our method produces detailed 3D face shapes in viewing conditions where existing state of the art often break down.*

## 1. Introduction

Estimating 3D face shapes from single images is a problem with a history now spanning two decades [54]. During this time, the related problem of recognizing faces in images has graduated to the point where modern systems can produce invariant and discriminative face representations for the most extreme face photos [35]. By comparison, 3D reconstruction methods have remained far behind.

Broadly speaking, existing face reconstruction methods are designed with either one of two goals. The first goal, exemplified by early 3D morphable models (3DMM) [1, 4, 36], some later shape from shading (SfS) techniques [27, 30], and several recent deep learning methods [39, 44], is obtaining highly detailed 3D face shapes. These methods produce discriminative results which are often detailed enough for subjects to be recognizable from their reconstructions. These results, however, are typically restricted to relatively easy viewing conditions. In particular, when faces are partially occluded, these methods often indiscriminately reconstruct the occlusion or fail completely.

Other face reconstruction methods were developed with invariance to viewing conditions as their goal. Some of these methods estimate face shapes by localizing facial landmarks [26, 48, 59]; others take an example based approach [15]. Deep methods were also recently proposed for this purpose [6, 11, 25, 51, 52]. All of these methods, however, sacrifice details in order to avoid failing on challenging, unconstrained photos. Their reconstructions typically provide few details [15, 48, 52] or are very generic [59].

Unlike previous work, we describe a method designed to attain both goals: *Detailed 3D face reconstruction and robustness to viewing conditions*, including, in particular, out of plane head rotations and occlusions (Fig. 1).

Our approach is inspired by the age old computer graphics concept of *bump mapping* [5] which involves separation of global shape from local details. When reconstructing faces, this implies estimating a global *foundation shape* separately from its local, *mid-level features* which are layered on the foundation. This modular design is very different from recent state of the art methods which explicitly couple global shape estimation with local details [44]. We see this separation as particularly important, as it allows estimating a robust global face shape, even under challenging conditions where estimating mid-level features breaks down.

To our knowledge, despite being intuitive, our design is novel. We additionally make the following technical contributions: (a) We propose deep estimation of facial bump maps in Sec. 4.1, (b) example based, bump map hole filling, for estimating plausible facial details in regions occluded by obstructions (Sec. 4.2), and (c) 3D soft symmetry for estimating self-occluded facial details in Sec. 4.3.

We report quantitative tests measuring the capabilities of these contributions (Sec. 5). We further provide qualitative comparisons of our reconstructions with recent state of the art methods. These results show our method to provide detailed reconstructions in extreme settings where previous results are either generic, low resolution, or else fail completely. Finally, code, deep models, and high resolution images are available on our project page.[1]

## 2. Related work

Before reviewing single view 3D reconstruction methods, we note that some methods were developed for *multi view* reconstruction [19, 28, 42, 43, 47]. With multiple views, better guarantees can be made on reconstruction accuracy and occlusion handling. We focus on single view settings where this information is not available.

**Reconstruction by example.** These techniques include some of the earliest methods for face reconstruction [16, 17, 54] and more recent approaches [15]. These methods use reference 3D face shapes to modify the shape estimated for an input face photo. Generally speaking, these methods were designed with an emphasis on robustness to unconstrained viewing conditions rather than fidelity or fine details. Note that like us, one of these methods actually produced shape estimates for occluded objects [16, 17]. By comparison, our method is designed to perform well in similar or even harder viewing conditions, including occlusions, yet provides detailed and accurate estimates.

**Face shape from facial landmarks.** Detected facial landmarks can be used to constrain 3D face shapes [26, 59]. Such methods often focus on landmark detection accuracy rather than facial details. As such, they are remarkably robust, yet produce generic 3D face shapes with few details. Furthermore, it is unclear how these methods would perform when presented with face images where landmarks are occluded or otherwise difficult to detect.

**SfS.** By making assumptions on the light sources and the reflectance properties of face, SfS methods showed accurate and often detailed reconstructions [27, 30]. The application of such methods is limited to scenes meeting their assumptions. Moreover, these methods typically reconstruct not only facial regions but also any objects occluding the faces.

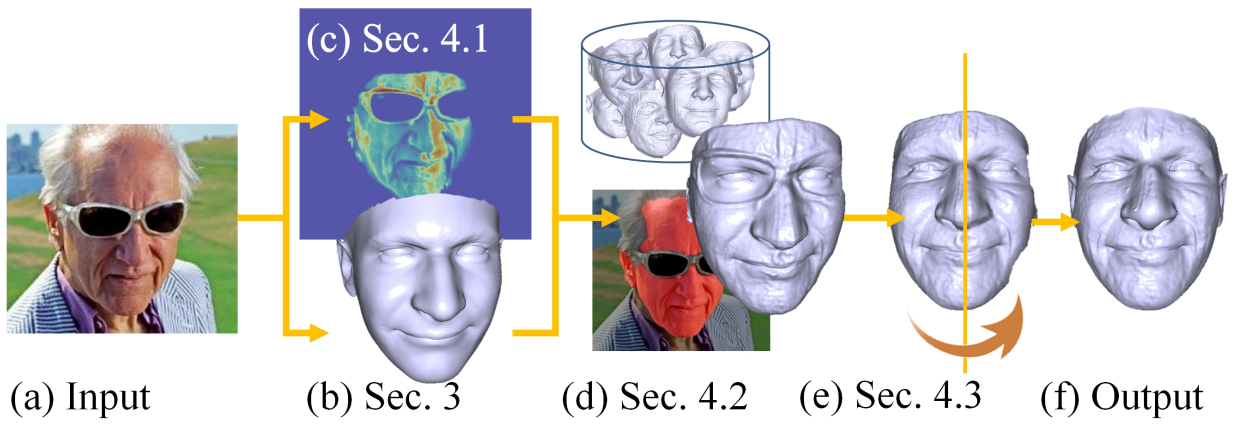

**Figure 2:** *Method overview.* See related sections for details.

**Statistical representations.** The most widely recognized examples in this category are the 3DMM fitting methods, originally proposed by Blanz and Vetter [3]. Since that seminal work, many improvements were made to the methods used to recover 3DMM face shape parameters [2, 10, 36, 40, 49, 55]. We rely on this representation and provide a short overview in Sec. 3. We fit 3DMM parameters very differently than the analysis by synthesis approach of these earlier methods, instead using a deep convolutional neural network (CNN) to regress 3DMM parameters and facial details directly from image intensities.

**Deep face shape estimation.** In keeping with the spirit of the times, deep networks have recently been applied to face shape estimation. Two main approaches were proposed. In one, rather shallow networks are trained on synthetically produced face shapes [38]. Facial details are added by training an end-to-end system to additionally estimate SfS [39]. To allow for detailed reconstructions unrestricted by the limitations of the 3DMM representation, face shapes were estimated directly using a depth map in [44].

Contrary to these methods, deep networks were used in [6, 11, 25, 45, 51, 50, 52] to estimate 3D shapes with an emphasis on unconstrained photos. These methods estimate shapes which are highly invariant to viewing conditions but provide only coarse 3D details.

We use deep networks to regress various elements of our face reconstructions. Our method, however, combines the benefits of both approaches, offering detailed reconstructions which are robust to viewing conditions.

## 3. Preliminaries: Laying the foundation

Our approach is motivated by this simple observation:

*3D face reconstruction involves the conflicting requirements of a strong regularization for a global shape vs. a weak regularization for capturing higher level details.*

End-to-end reconstruction systems can easily fail to balance these two requirements. Indeed, as we later show (Sec. 5.3), existing methods are often either over-regularized (sacrific-ing fine details for invariance) or under-regularized (sacrificing invariance for details).

[1] Available: github.com/anhttran/extreme_3d_faces

To effectively combine these two requirements, we instead take a modular approach (see Fig. 2). Our method is inspired by computer graphics, bump map representations [5] which separate global shape from local details. This separation allows for strong regularization over the global face shape – thereby providing robustness even in the most challenging viewing conditions – and weak regularization over the local details. Thus, failure to estimate details does corrupt the final result; occluded details can be estimated without modifying the global shape.

Given an input face photo $\mathbf{I}$, we separately estimate the following shape components: A foundation shape, $\mathbf{s}$, facial expression $\mathbf{e}$, and 6 degrees of freedom (6DoF) viewpoint, $\mathbf{v}$, all described below. Next, we estimate a bump map $\boldsymbol{\Delta}$, which captures facial wrinkles and other non-parametric, mid-level features (Sec. 4.1). Finally, we complete missing facial details due to occlusions to produce our output 3D shape (Sec. 4.2 and 4.3).

**The foundation shape.** We use the standard, linear 3DMM representation of 3D faces [1, 4, 10, 20, 36]. We refer to any of these sources for details on this representation. Briefly, we model a face as:

$$\mathbf{s} = \widehat{\mathbf{s}} + \sum_{i=1}^{s} \boldsymbol{\alpha}_i \mathbf{W}_i^S, \tag{1}$$

where $\widehat{\mathbf{s}}$ represents an average 3D face shape, $\boldsymbol{\alpha} \in \mathbb{R}^s$ are subject-specific, face shape coefficients estimated from $\mathbf{I}$, and $\mathbf{W}^S \in \mathbb{R}^{3n \times s}$ are principal components representing the distribution of 3D shapes. Here, $3n$ represents the 3D coordinates for $n$ vertices of our 3D faces. We use the Basel face model (BFM) [36], which provides both $\widehat{\mathbf{s}}$ and $\mathbf{W}^S$, and defines $s = 99$ as the number of shape components.

Given $\mathbf{I}$, we estimate $\boldsymbol{\alpha}$ (and hence, $\mathbf{s}$) using the recent deep 3DMM approach of Tran *et al.* [52], using their publicly available code and pre-trained model. They regress 3DMM coefficients, $\boldsymbol{\alpha}$, directly from image intensities using a ResNet architecture with 101 layers [18].

**Estimating facial expressions.** We model expressions similarly to shape, using the following formulation:

$$\mathbf{e} = \sum_{j=1}^{m} \boldsymbol{\eta}_j \mathbf{W}_j^E. \tag{2}$$

Expression deformations are represented as linear combination of expression coefficients $\boldsymbol{\eta} \in \mathbb{R}^m$ (estimated from $\mathbf{I}$) and expression components which span the space of deformations, $\mathbf{W}^E \in \mathbb{R}^{3n \times m}$. Here again, $3n$ represents the 3D coordinates for the $n$ vertices of BFM. We use the $m = 29$ expression components provided by 3DDFA [59].

Expressions were estimated using the same system provided by the authors of [52], though a more robust method would use the recent, deep ExpNet [9]. Comparing expression estimation methods on occluded faces is an ill-defined task: After all, whenever parts of the face are occluded, the expression coefficients controlling occluded facial regions can be arbitrary. Our decision to use this method is therefore motivated simply by the utility of their public code.

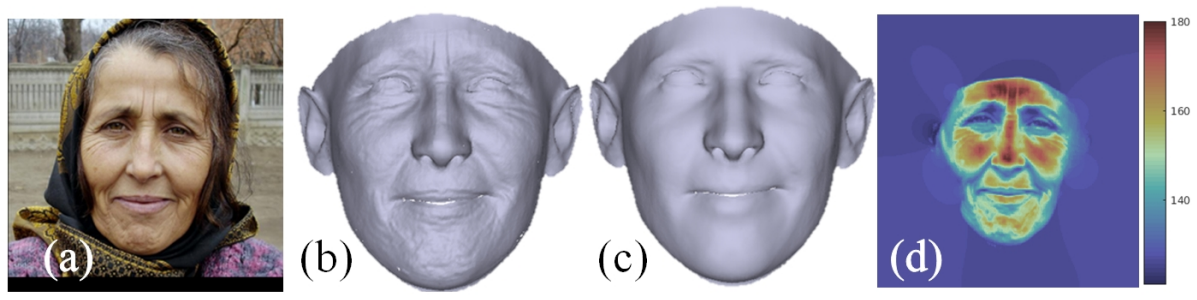

**Figure 3:** *Illustrating Eq.* (3). For an input image (a), the difference between the depths of a detailed shape (b) and foundation shape (c) provides our bump map (d) (visualized here as a heat map). See text for more details.

Note that the shape from Eq. (1), $\mathbf{s}$, and the expression $\mathbf{e}$ from Eq. (2) can be summed, obtaining $\mathbf{F} = \mathbf{s} + \mathbf{e}$, representing an expression adjusted foundation shape.

**Estimating viewpoint.** We represent viewpoint as $\mathbf{v} = [\mathbf{r}^T, \mathbf{t}^T]$, where $\mathbf{r} \in \mathbb{R}^3$ is the 3D rotation, represented in Euler angles, and $\mathbf{t} \in \mathbb{R}^3$, a translation vector, together aligning a generic 3D face shape with the face appearing in $\mathbf{I}$. Given $\mathbf{v}$, a 3D to 2D projection matrix can be estimated using standard means [14], projecting the surface points of our 3D shape onto the input image. To estimate $\mathbf{v}$ given $\mathbf{I}$, we use the deep, FacePoseNet method proposed in [8]. Fig. 2 (d) provides an example foundation shape, modified for expression and aligned with input $\mathbf{I}$.

## 4. Adding mid-level details

Given an expression adjusted foundation shape, we layer it with mid-level features estimated directly from $\mathbf{I}$. Importantly, rather than estimating a detailed shape directly [44], we estimate *local deformations of the shape*. Our foundations shape thus provides strong regularization over the global shape and allows facial details to be estimated without sacrificing robustness to extreme viewing conditions. As we later show, this separation also allows us to predict missing details in places where the face is occluded from view, by estimating the local deformations in those regions.

### 4.1. Image to bump map translation

Inspired by [7], we model local deformations of the face surface as *depth displacements*. That is, as displacements of the *depth map* defined by measuring the distances along the rays emanating from the center of projection, through the pixels of $\mathbf{I}$ to the face surface. (See Fig. 2 (c-d) and 3).

Formally, we store bump map $\boldsymbol{\Delta}(\mathbf{p})$ in a matrix of the same spatial dimensions as $\mathbf{I}$, simply defined as:

$$\boldsymbol{\Delta}(\mathbf{p}) = \begin{cases} \theta\big(z'(\mathbf{p}) - z(\mathbf{p})\big) & \text{face projects to } \mathbf{p} \\ \theta(0) & \text{all other pixels} \end{cases}, \tag{3}$$

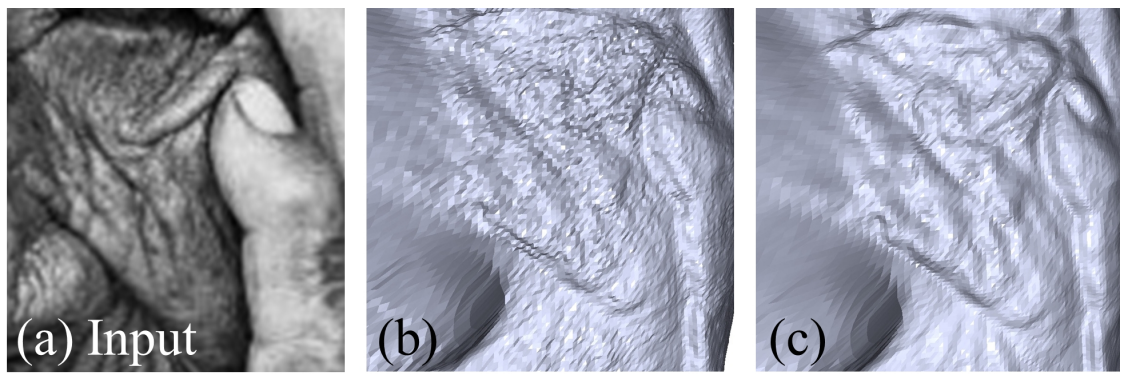

**Figure 4:** *Effect of our loss from Eq.* (5). (a) Close-up view of an input $\mathbf{I}$. (b) Surface detailed by a network trained with standard $\ell_1$ loss. (c) Surface details obtained following training with our modified loss (Eq. (5)) are less noisy, without compromising high frequencies.

where $\mathbf{p} = [x, y]$ is a pixel coordinate in $\mathbf{I}$, $z'(\mathbf{p})$ is the distance along the viewing ray from the surface of the detailed face shape to $\mathbf{p}$ (its *depth*), $z(\mathbf{p})$ is the depth of the foundation shape at $\mathbf{p}$, and $\theta(\cdot)$ is a linear encoding function, here, bringing depth values to the range $[0, ..., 255]$.

The *bump* we aim to compute at every pixel $\mathbf{p}$ is thus $\delta = z'(p) - z(p)$. Given a bump map $\mathbf{\Delta}$ and the depth of the foundation, we can easily compute a detailed depth by:

$$z'(\mathbf{p}) = z(\mathbf{p}) + \theta^{-1}(\mathbf{\Delta}(\mathbf{p})). \tag{4}$$

Note that this detailed depth corresponds to a dense 3D face model $\mathbf{D}$, where each face pixel in the depth map defines a 3D point on the surface of the detailed face shape.

**Bump map training set.** We produce bump maps using a deep encoder-decoder framework [24]. Training such a network requires a substantial number of face images, all assigned with target bump maps. To obtain such a set, we estimate bump maps using an existing method applied to a large face image set. These estimates are then used as surrogates for ground truth.

Specifically, we used face photos from the VGG set [35]. We estimate a foundation shape along with 6DoF viewpoint for each image (Sec. 3). The viewpoint was used to project the estimated 3D shape onto the image [14], providing depth values, $z(\mathbf{p})$, along with a face / background mask. We then applied a state of the art SfS method [34] to the face (foreground) region in each image. The detailed depth estimates produced by this method are then taken to be $z'(\mathbf{p})$. With both foundation and detailed depths, we use Eq. (3) to compute a bump map for each image (See Fig. 3).

The SfS method we used is rather slow, taking minutes to compute a single depth map. This limited application of the process to a subset of the VGG set. We further manually selected only images where SfS produced reasonable depth estimates (determined by visual inspection). Consequently, we were left with 4,300 images assigned with target bump maps. Of these, 4,200 were used to train our method and the reminder as a validation set.

**Learning to estimate bump maps.** We estimate bump maps with a framework similar to the one used for image translation [24] with some notable modifications. Specifically, since our goal emphasizes facial details, we increase both the input and the output resolution of this network to 500×500. This also implies a deeper network than the ones originally used in [24]: we use seven encoding blocks and six decoding blocks. Finally, as reported by others, this approach tends to produce output with checkerboard artifacts. To mitigate this, we use *resize and convolutions* rather than transposed convolutions [33].

Importantly, we define our own network loss, which we found to suppress noise without sacrificing high frequency details, as follows:

$$\ell_{bump} \doteq \ell_1\Big(\widetilde{\mathbf{\Delta}} - \mathbf{\Delta}\Big) + \\ + \ell_1\Big(\frac{\partial\widetilde{\mathbf{\Delta}}}{\partial x} - \frac{\partial\mathbf{\Delta}}{\partial x}\Big) + \ell_1\Big(\frac{\partial\widetilde{\mathbf{\Delta}}}{\partial y} - \frac{\partial\mathbf{\Delta}}{\partial y}\Big). \tag{5}$$

Here, $\ell_1 = ||\cdot||_1$ is the classic L1 loss which does not tend to over-smooth results. The 2D gradient of the bump map is expressed by $\frac{\partial\mathbf{\Delta}}{\partial x}, \frac{\partial\mathbf{\Delta}}{\partial y}$. We found that by adding these last two terms we reduce bump map noise by favoring smoother surfaces, but still allowing high frequency details to be preserved. A qualitative effect of the new loss proposed in Eq. (5) can be appreciated in Fig. 4. At test time, this trained deep encoder-decoder is used to translate an input RGB image $\mathbf{I}$ into a bump map $\mathbf{\Delta}$.

## 4.2. Recovering occluded details

Similarly to other methods for detailed face reconstruction [15], our bump map estimation method of Sec. 4.1 is not invariant to occlusions. Glasses, for example, are often reconstructed along with the surface of the face (Fig. 2 (d)). By separately reconstructing a foundation shape invariant to these obstructions (see Sec. 5.1) and mid-level details represented as bump maps, these errors can be corrected by modifying the 2D bump map directly using techniques similar to those used for image inpainting.

Specifically, we use an existing face segmentation method [32] to determine occluded regions in the input image $\mathbf{I}$. Because $\mathbf{I}$ and its bump map, $\mathbf{\Delta}$, are both aligned, the segmentation obtained on $\mathbf{I}$ can be directly applied to $\mathbf{\Delta}$. We consider any non-face regions as holes which we fill using an example based technique.

**Example based hole filling approach.** We use a collection of reference bump maps from which we borrow unoccluded details to complete missing regions of our bump map. Given image $\mathbf{I}$ we search through this reference set for one presenting a suitably similar individual. We then transfer details from the bump map associated with the selected reference image to the occluded regions in $\mathbf{\Delta}$.

We use deep features designed for face recognition to encode identity and to search for matching reference faces.

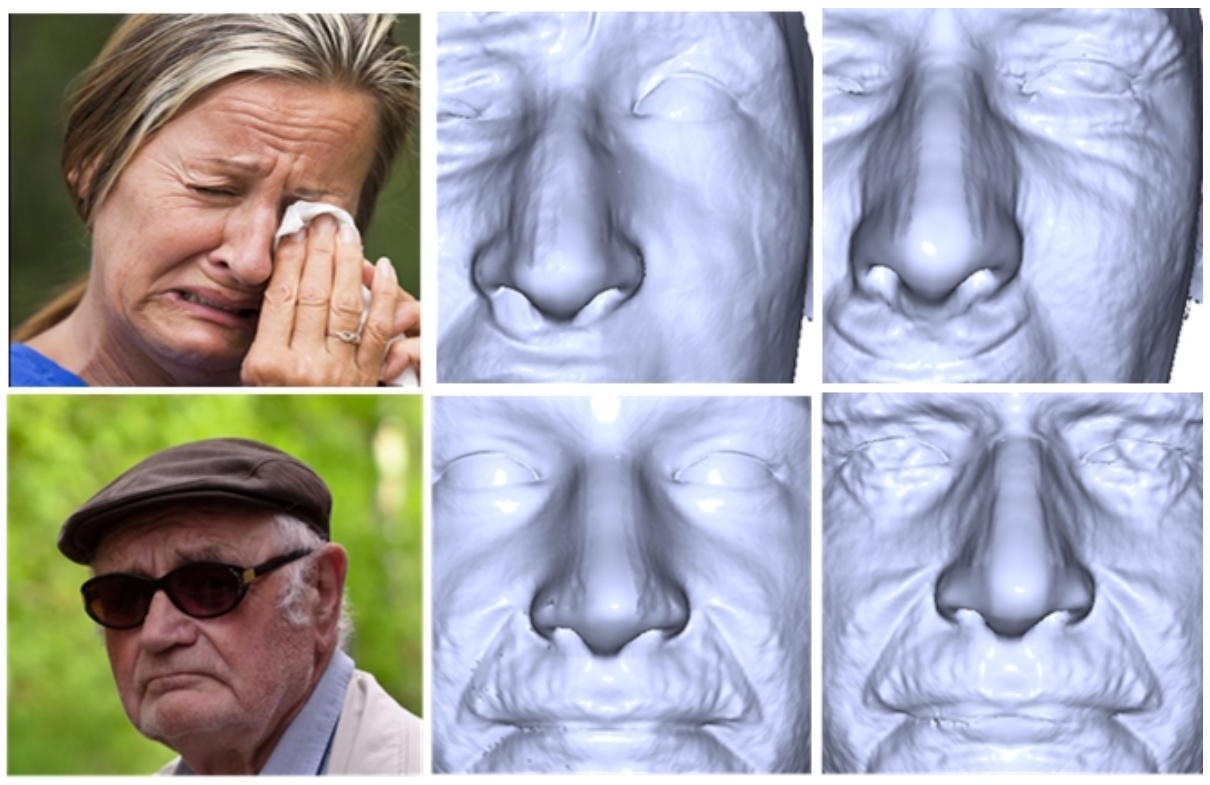

**Figure 5:** *Hole filling comparison.* 3D Reconstruction with occlusions, comparing Yeh *et al.* [56] with our bump map hole filling (Sec. 4.2). Their $64 \times 64$ resolution limit over-smooths results compared to our example based method.

Our rationale is simple: Deep features produced by modern face recognition systems are designed to be highly robust to appearance variations and particularly occlusions. We therefore expect them to match the partially occluded input face with a reference image of a similar individual (with similar facial details) regardless of occlusions.

**Searching the reference set.** Our references consist of 12K unoccluded face images selected from the CASIA [57] and VGG [35] sets. We search this set for $k = 100$ images with viewpoints most similar to the one in the input ($\mathbf{v}$). By ensuring that both reference and input share similar poses we mostly eliminate the need to address pose differences when transferring details between their bump maps.

From this short list, we select the face with the most similar deep features. For this purpose, we use the publicly available deep models trained for face recognition by [31]. Deep feature similarity is determined using the cosine distance.

**Blending details.** We transfer details from the bump map $\mathbf{\Delta}_{r^*}$ associated with the selected reference, to fill-in holes in $\mathbf{\Delta}$. To this end, we align the two faces in 2D using standard, similarity transform alignment. With the two bump maps registered, we use the occlusion mask to update only occlusions in $\mathbf{\Delta}$. Details transferred over from $\mathbf{\Delta}_{r^*}$ onto these regions are blended into their surroundings using gradient-based blending [37].

**More sophisticated inpainting?** Many recent methods were proposed for filling-in holes in images with some recent ones including [22, 56]. Though we tried several alternatives, our best results were produced with the method described above. This is demonstrated in a comparison with the recent, state of the art method of Yeh *et al.* [56], in Fig. 5. Their system is limited to $64 \times 64$ pixel input images. This low resolution is apparent in the over smooth results produced by their method compared to our example based approach which is less restricted by scale and can handle far larger images.

As our tests in Sec. 5.2 show, our hole filling is not only more plausible but actually recovers much of the original discriminative details which were lost due to the occlusion.

### 4.3. Soft symmetry based model completion

Our bump map estimation does not provide features for regions of the face turned away from the camera. Once details are estimated for regions occluded due to obstructions (Sec. 4.2), we therefore proceed to estimate details missing due to self occlusions. We make the standard assumption of face symmetry [12] and propose to transfer bump map details from visible face regions to self occluded regions, blending them smoothly to produce the result.

BFM, used to represent the foundation shape, provides a complete face representation and their vertices allow for easy indexing of symmetric face regions. These properties are very useful for transferring details between symmetric face regions. Our bump maps, however, are produced in the same resolution as the input $\mathbf{I}$. This resolution is typically higher than the one defined by BFM for the density of its 3D vertex distribution. To effectively recover such high resolution details in the occluded region, we use soft symmetry applied to the BFM (despite being sparse and low resolution) as an intermediate representation.

**Dense to sparse conversion.** We first convert our dense 3D model $\mathbf{D}$ (Sec. 4.1), to a sparse mesh, $\mathbf{R}$, represented using BFM. This process is simple: we project each visible vertex of the foundation shape, $\mathbf{F}$, onto the detailed depth map, acquire the new depth, and update its 3D coordinates. Any occluded vertices will be updated later by soft symmetry.

**Dense from sparse registration.** In order to recover a dense 3D model from the sparse one, we register each 3D vertex of the dense mesh $\mathbf{D}$ – corresponding to a pixel in the detailed depth map – to its corresponding triangle on the sparse mesh $\mathbf{R}$. Specifically, let $\mathbf{p}^D$ be a pixel in the 2D depth map and $\mathbf{P}^D$ its corresponding vertex in a dense (high resolution) 3D face. A triangle in the sparse representation is defined by its three vertices and their corresponding three projections onto the depth map: $\{(\mathbf{P}_i^R, \mathbf{p}_i^R)|i = 1, ..., 3\}$.

A correspondence between the dense pixel location $\mathbf{P}^D$ and its three sparse counterparts, is computed using the 2D alignment parameters $u, v \in [0, 1]$ such that:

$$\mathbf{p}^D = u\mathbf{p}_1^R + v\mathbf{p}_2^R + (1 - u - v)\mathbf{p}_3^R. \tag{6}$$

The residual in 3D, is then defined as:

$$\delta\mathbf{P} = \mathbf{P}^D - (u\mathbf{P}_1^R + v\mathbf{P}_2^R + (1 - u - v)\mathbf{P}_3^R). \tag{7}$$

Finally, we store the parameter set $(u, v, \delta\mathbf{P})$ in order to recover $\mathbf{P}^D$ from $\{\mathbf{P}_i^R\}$:

$$\mathbf{P}^D = (u\mathbf{P}_1^R + v\mathbf{P}_2^R + (1 - u - v)\mathbf{P}_3^R) + \delta\mathbf{P}. \tag{8}$$

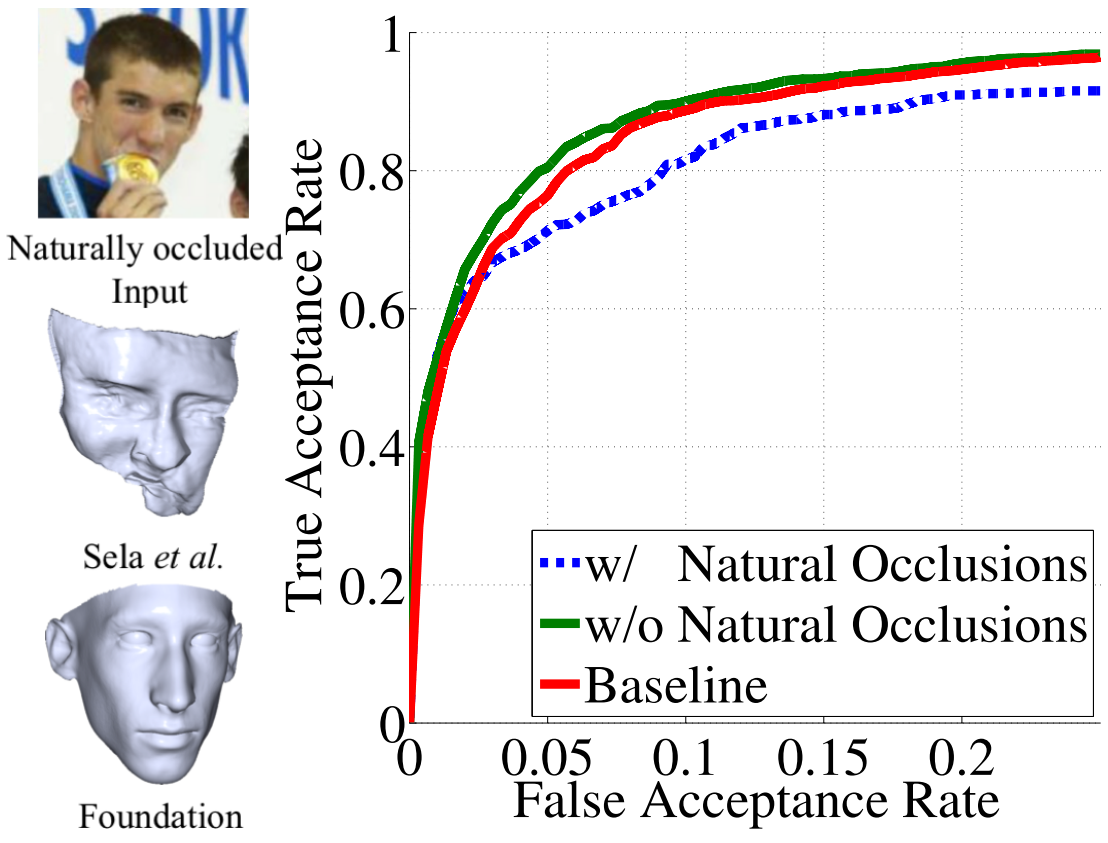

**Figure 6:** *Reconstructions with occlusions.* Left: Qualitative results of Sela *et al.* [44] and the method used as our foundation [52]. The loose regularization required by [44] to produce fine details results in a corrupt shape for the occluded face. The foundation sacrifices details but remains robust. Right: LFW verification ROC for foundation shapes [52], with and without occlusions.

**Soft symmetry on the sparse mesh.** Self occluded face regions are defined as those where the normals of the 3D face shape, point away from the image plane. For every such point $\mathbf{P}_i$ on the sparse representation $\mathbf{R}$, we locate its corresponding point, $\mathbf{P}_i^f$, by reflection over the vertical axis of symmetry. We then use Poisson blending to recover this occluded point [37, 58]:

$$\nabla_R \mathbf{P}_i = \begin{cases} \nabla_R \mathbf{P}_i^f & \text{if } \mathbf{P}_i^f \text{ is visible} \\ \nabla_F \mathbf{P}_i^I & \text{otherwise,} \end{cases} \tag{9}$$

where $\nabla_R$ and $\nabla_F$ are the discrete graph Laplacians on $\mathbf{R}$ and $\mathbf{F} = \mathbf{s} + \mathbf{e}$, respectively.

**Soft symmetry on the dense mesh.** We now have a complete and detailed sparse mesh $\mathbf{R}$. The 3D details on the self-occluded area can be recovered by converting this mesh back to the dense structure using symmetry-based inference, as follows.

For each point $\mathbf{P}^D$ on $\mathbf{D}$, we already registered it to a triangle $\{\mathbf{P}_i^R | i = 1, ..., 3\}$ on $\mathbf{R}$ with an alignment parameter set $(u, v, \delta P)$ using Eq. (8). We denoted the opposite triangle on $\mathbf{R}$, through the axis of symmetry, as $\{\mathbf{P}_i^{Rf} | i = 1, ..., 3\}$. If any point in $\{\mathbf{P}_i^{Rf}\}$ is newly recovered from Poisson blending, we can define the reflection of $\mathbf{P}^D$ over the symmetry axis, denoted $\mathbf{P}^{Df}$, as follows:

$$\mathbf{P}^{Df} = (u\mathbf{P}_1^{Rf} + v\mathbf{P}_2^{Rf} + (1 - u - v)\mathbf{P}_3^{Rf}) + \delta \mathbf{P}^f, \tag{10}$$

where $\delta \mathbf{P}^f$ is the reflection of $\delta \mathbf{P}$ over the symmetry axis.

**Producing the final shape.** Symmetry-based inference cannot fill all missing regions in $\mathbf{D}$; sometimes occlusions appear on both sides (e.g. chin, nose, lower jawline). Therefore, the dense and sparse mesh are combined into the final mesh $\mathbf{S}$, which is both complete and dense. We first remove the overlapping regions from $\mathbf{R}$, then *zipper* the meshes using the well-known technique of [53].

| Method | 100%-EER | Accuracy | nAUC |
|---|---|---|---|
| Foundation s and occlusions (See Sec. 5.1, and Fig. 6) | | | |
| Tran *et al.* [52] | 89.40±1.52 | 89.36±1.25 | 95.90±0.95 |
| w/ Occ. | 86.96±1.70 | 87.86±1.35 | 94.09±0.86 |
| w/o Occ. | 90.30±0.83 | 89.91±0.86 | 96.46±0.44 |
| Bump maps $\mathbf{\Delta}$ and hole filling (See Sec. 5.2 and Fig. 7) | | | |
| Baseline (bump maps) | 92.76±1.34 | 92.76±1.22 | 98.17±0.63 |
| w/ Occ. | 75.66±2.00 | 75.96±1.98 | 83.73±1.86 |
| Ex. based filling | 83.87±1.79 | 84.06±1.96 | 91.87±1.43 |

**Table 1:** *Quantitative evaluations on LFW.*

# 5. Results

Our foundation shape, its expression, and viewpoint are estimated using publicly available code (Sec. 3). These steps require ∼0.3s per image. Deep bump-map regression is implemented in PyTorch, and requires a further 0.03s/image. Segmenting faces requires 0.02s/image. To search for suitable examples, transfer, and blend their details into occluded regions of our bump map requires 0.6s/image (Sec. 4.2). Our unoptimized C++ implementation for soft symmetry (Sec. 4.3) currently takes ∼50s/image. These runtimes were measured on a desktop system with an Intel(C) Xeon(R) CPU X5687 @ 3.60GHz 4, 12GB RAM, and GeForce GTX 1080.

## 5.1. Occlusion invariance of the foundation shape

Any number of 3DMM fitting methods could potentially be used to estimate a foundation shape in Sec. 3. Our choice of using deep 3DMM [52] is motivated by the quantitative results provided in that paper, demonstrating its unique robustness to extreme viewing conditions. To fully support its application to occluded face images, however, we extend their tests by evaluating their method on occluded faces.

**Test settings and verification system.** To this end, we use images from the Labeled Faces in the Wild (LFW) benchmark [21]. We automatically select partially occluded LFW images, using a recent, state of the art face segmentation method [32] (Fig. 6 (top left)). We then conducted face verification tests using only pairs where at least one of the faces was occluded, where both faces were unoccluded, and the original LFW set. We used the same face verification system from [52], and we refer to that paper for more details.

**Occlusion invariance results.** Fig. 6 (left) shows the sensitivity of the state of the art method of [44] to these occlusions. Their failure is likely due to their weak regularization over the global face shape, required to produce fine details. The figure also visualizes the shape estimated by [52]. Its much stronger regularization provides a natural face shape despite any occlusion. Though this regularization also limits the details produced by that method, we use it only as a foundation and add details separately.

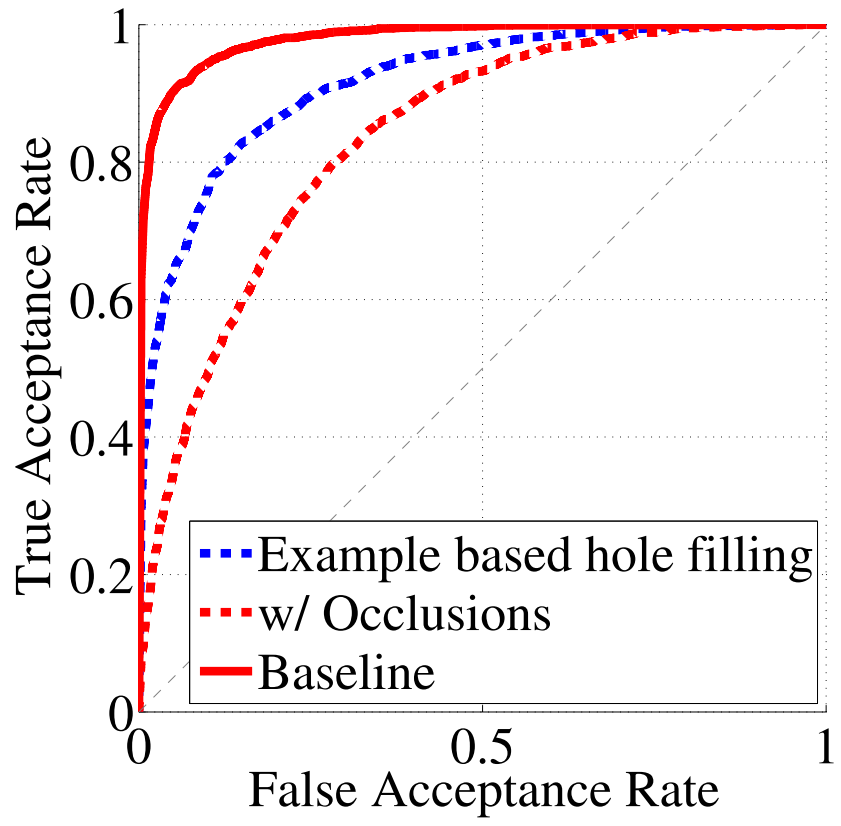

**Figure 7:** *LFW verification on bump maps.* Our example based hole filling clearly restores identifiable features to occluded bump maps (dashed blue vs. dashed red).

We further quantitatively verify the robustness of [52] to occlusions. Tab. 1 (top) reports verification results on the LFW benchmark with and without occlusions, using the foundation shape as a representation (see also ROC in Fig. 6 (right)). Though occlusions clearly impact recognition, this drop is limited, demonstrating the robustness of [52] to occlusions and supporting our decision to use this method for our foundation shapes.

## 5.2. Bump map hole filling and identifiable features

Our bump map hole filling method (Sec. 4.2) is designed to produce only plausible details. Still, it is interesting to consider how well this process correctly estimates discriminative features. To this end, we again use the LFW benchmark, this time measuring verification rates on our estimated bump maps, rather than the original images.

Specifically, we measure accuracy on (a) bump maps estimated from LFW images, (b) bump maps with occluded regions, and (c) bump maps with occlusions completed using the example based method of Sec. 4.2. The first tests measure how well bump maps capture identities. The second measure the impact of occlusions on these representations. The third measure if and how hole filling affects the identity perceived from the bump maps.

**Test settings and verification system.** We randomly position occlusions (black squares) of $75{\times}75$ pixels on bump maps estimated for LFW images. Bump maps were represented as 2D images, with bumps in the range of [0,...,255] (Sec. 4.1). We used a simple recognition system, favoring it over more elaborate ones to emphasize the effect of the information reflected by the bump maps rather than the method used to recognize them. This system encoded bump maps using a standard AlexNet [29], trained on bump maps extracted from the entire CASIA set [57]. Following the current best practice we added Batch Normalization layers [23] to ease the training and removed dropout. We used transfer learning from weights pre-trained on ImageNet; further details can be found in [46].

**Hole filling results.** Our results are presented in Fig. 7 and offer several noteworthy conclusions. First, the original bump maps (solid red) capture identifiable features; recognition results on the bump maps before holes were introduced are very high. This is not surprising: bump maps essentially reflect SfS which is known to be discriminative. Also, not surprising is that by introducing occlusions, these recognition rates drop substantially (dashed red). Remarkably, however, our example based hole filling *recovers much of this drop in accuracy* (dashed blue). These results imply that not only does our example based hole filling produce plausible estimates for occluded regions, these estimates are a *good approximation of the true, occluded details*.

## 5.3. Qualitative results

Rendered views of reconstructed 3D faces produced by our method are provided throughout the paper. A comparison of our results to those produced by recent state of the art is provided in Fig. 8. Baseline methods include the analysis by synthesis 3DMM fitting of [41], the flow based approach of [15], the recent 3DDFA method for facial landmark detection and shape estimation [59], the deep 3DMM method used to produce our foundation shapes [52], and the deep system for detailed reconstructions of [44].

The recent method of Sela *et al.* produces beautifully detailed results when the face is relatively unoccluded (Fig. 8 (c)), but due to its weaker regularization over the global shape, fails completely on occluded faces and other challenging views. Other methods are either not regularized enough (3DMM and flow, in rows (a) and (e)) or are too regulated, producing generic shapes (3DDFA).

# 6. Conclusions

We describe a method capable of producing detailed 3D face reconstructions from photos taken in extreme viewing conditions. Our method goes on to estimate plausible facial details in places where the face is occluded. Our results represent a leap in 3D face reconstruction capabilities, previously confined to mostly frontal, unoccluded views or to 3D shapes with limited details. At the heart of our approach is its modular design, which decouples the task of estimating a robust foundation shape from the task of estimating its mid-level details, represented here as bump maps. This approach goes against the grain of more recent methods, which explicitly seek to train deep, end-to-end systems for this purpose. Although there is no denying the benefits of such end-to-end systems, this paper shows the potential ad-vantage modular designs can sometimes have. This advantage is particularly relevant for faces, a class where domain knowledge is abundant and object shapes are well regulated.

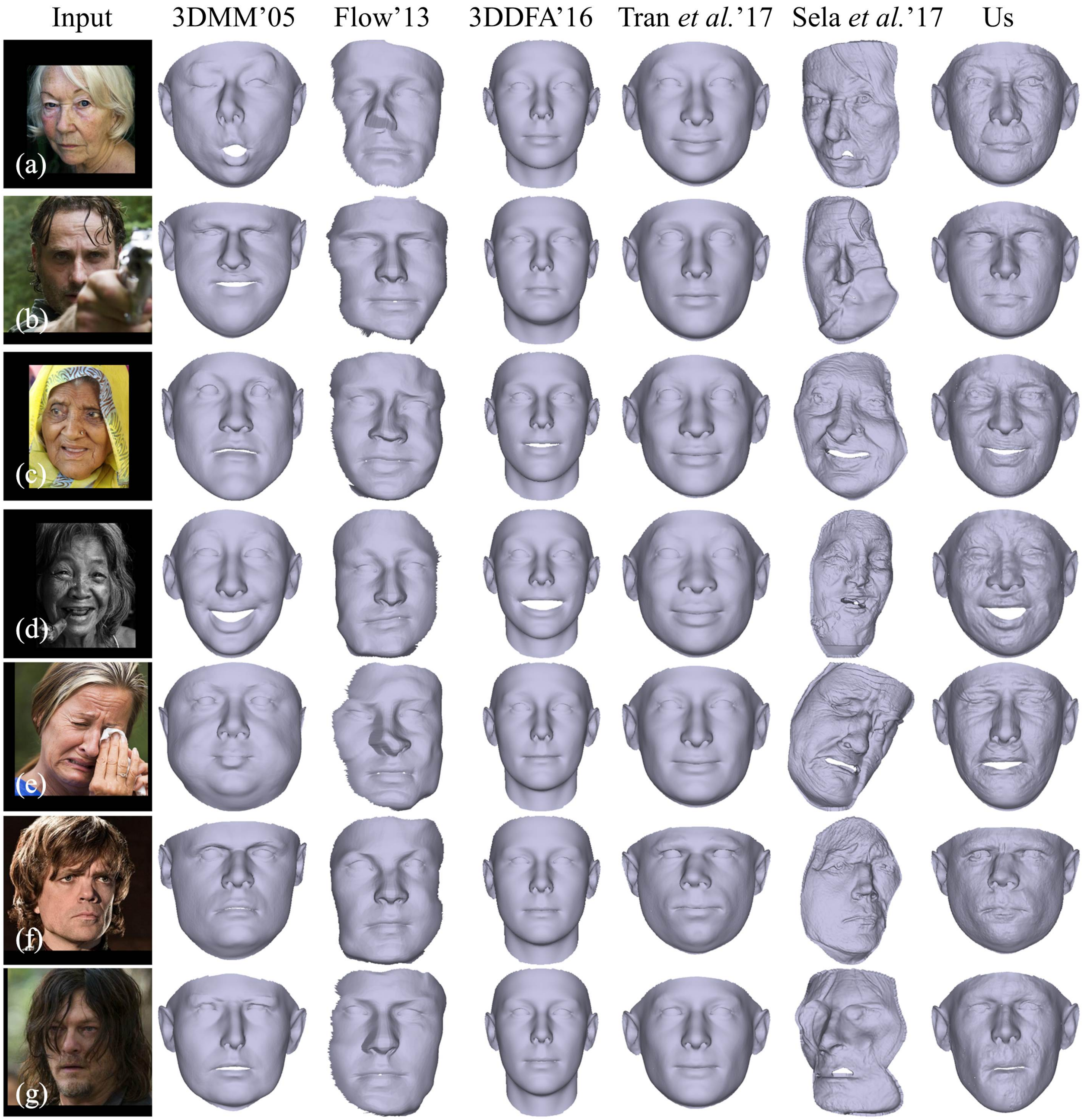

**Figure 8:** *Comparison of qualitative results.* Baseline methods from left to right: 3DMM fitting of [41], flow based approach [15], 3DDFA [59], Tran *et al.* [52] (used to produce our foundations), and the end-to-end system of Sela *et al.* [44].

## Acknowledgments

This research is based upon work supported in part by the Office of the Director of National Intelligence (ODNI), Intelligence Advanced Research Projects Activity (IARPA), via IARPA 2014-14071600011. The views and conclusions contained herein are those of the authors and should not be interpreted as necessarily representing the official policies or endorsements, either expressed or implied, of ODNI, IARPA, or the U.S. Government. The U.S. Government is authorized to reproduce and distribute reprints for Governmental purpose notwithstanding any copyright annotation thereon.

## References

[1] V. Blanz, S. Romdhani, and T. Vetter. Face identification across different poses and illuminations with a 3d morphable model. In *Int. Conf. on Automatic Face and Gesture Recognition*, pages 192–197, 2002. 1, 3

[2] V. Blanz, K. Scherbaum, T. Vetter, and H.-P. Seidel. Exchanging faces in images. *Comput. Graphics Forum*, 23(3):669–676, 2004. 2

[3] V. Blanz and T. Vetter. Morphable model for the synthesis of 3D faces. In *Proc. ACM SIGGRAPH Conf. Comput. Graphics*, 1999. 2

[4] V. Blanz and T. Vetter. Face recognition based on fitting a 3d morphable model. *Trans. Pattern Anal. Mach. Intell.*, 25(9):1063–1074, 2003. 1, 3

[5] J. F. Blinn. Simulation of wrinkled surfaces. *Proc. ACM SIGGRAPH Conf. Comput. Graphics*, 12(3):286–292, 1978. 1, 3

[6] J. Booth, E. Antonakos, S. Ploumpis, G. Trigeorgis, Y. Panagakis, and S. Zafeiriou. 3D face morphable models "in-the-wild". In *Proc. Conf. Comput. Vision Pattern Recognition*, 2017. 1, 2

[7] C. Cao, D. Bradley, K. Zhou, and T. Beeler. Real-time high-fidelity facial performance capture. *ACM Trans. on Graphics*, 34(4):46, 2015. 3

[8] F. Chang, A. Tran, T. Hassner, I. Masi, R. Nevatia, and G. Medioni. FacePoseNet: Making a case for landmark-free face alignment. In *Proc. Int. Conf. Comput. Vision Workshops*, 2017. Available: https://github.com/fengju514/Face-Pose-Net. 3

[9] F. Chang, A. Tran, T. Hassner, I. Masi, R. Nevatia, and G. Medioni. ExpNet: Landmark-free, deep, 3D facial expressions. In *Int. Conf. on Automatic Face and Gesture Recognition*, 2018. Available https://github.com/fengju514/Expression-Net. 3

[10] B. Chu, S. Romdhani, and L. Chen. 3D-aided face recognition robust to expression and pose variations. In *Proc. Conf. Comput. Vision Pattern Recognition*, 2014. 2, 3

[11] P. Dou, S. K. Shah, and I. A. Kakadiaris. End-to-end 3D face reconstruction with deep neural networks. In *Proc. Conf. Comput. Vision Pattern Recognition*, July 2017. 1, 2

[12] R. Dovgard and R. Basri. Statistical symmetric shape from shading for 3D structure recovery of faces. *European Conf. Comput. Vision*, pages 99–113, 2004. 5

[13] X. Glorot and Y. Bengio. Understanding the difficulty of training deep feedforward neural networks. In *Proc. Int. Conf. on Artificial Intelligence and Statistics*, pages 249–256, 2010. 11

[14] R. Hartley and A. Zisserman. *Multiple view geometry in computer vision*. Cambridge university press, 2003. 3, 4

[15] T. Hassner. Viewing real-world faces in 3D. In *Proc. Int. Conf. Comput. Vision*, pages 3607–3614. IEEE, 2013. Available: www.openu.ac.il/home/hassner/projects/poses. 1, 2, 4, 7, 8, 14

[16] T. Hassner and R. Basri. Example based 3D reconstruction from single 2D images. In *Proc. Conf. Comput. Vision Pattern Recognition Workshops*. IEEE, 2006. 2

[17] T. Hassner and R. Basri. Single view depth estimation from examples. *arXiv preprint arXiv:1304.3915*, 2013. 2

[18] K. He, X. Zhang, S. Ren, and J. Sun. Deep residual learning for image recognition. In *Proc. Conf. Comput. Vision Pattern Recognition*, June 2016. 3

[19] M. Hernandez, T. Hassner, J. Choi, and G. Medioni. Accurate 3D face reconstruction via prior constrained structure from motion. *Computers & Graphics*, 2017. 2

[20] G. Hu, F. Yan, C.-H. Chan, W. Deng, W. Christmas, J. Kittler, and N. M. Robertson. Face recognition using a unified 3D morphable model. In *European Conf. Comput. Vision*, pages 73–89. Springer, 2016. 3

[21] G. B. Huang, M. Ramesh, T. Berg, and E. Learned-Miller. Labeled faces in the wild: A database for studying face recognition in unconstrained environments. Technical Report 07-49, UMass, Amherst, October 2007. 6

[22] S. Iizuka, E. Simo-Serra, and H. Ishikawa. Globally and locally consistent image completion. *ACM Trans. on Graphics*, 36(4):107, 2017. 5

[23] S. Ioffe and C. Szegedy. Batch normalization: Accelerating deep network training by reducing internal covariate shift. In *Int. Conf. Mach. Learning*, pages 448–456, 2015. 7

[24] P. Isola, J.-Y. Zhu, T. Zhou, and A. A. Efros. Image-to-image translation with conditional adversarial networks. In *Proc. Conf. Comput. Vision Pattern Recognition*, July 2017. 4, 11

[25] A. S. Jackson, A. Bulat, V. Argyriou, and G. Tzimiropoulos. Large pose 3D face reconstruction from a single image via direct volumetric CNN regression. *Proc. Int. Conf. Comput. Vision*, 2017. 1, 2

[26] A. Jourabloo and X. Liu. Large-pose face alignment via cnn-based dense 3D model fitting. In *Proc. Conf. Comput. Vision Pattern Recognition*, 2016. 1, 2

[27] I. Kemelmacher-Shlizerman and R. Basri. 3D face reconstruction from a single image using a single reference face shape. *Trans. Pattern Anal. Mach. Intell.*, 33(2):394–405, 2011. 1, 2

[28] I. Kemelmacher-Shlizerman and S. M. Seitz. Face reconstruction in the wild. In *Proc. Int. Conf. Comput. Vision*, pages 1746–1753. IEEE, 2011. 2

[29] A. Krizhevsky, I. Sutskever, and G. E. Hinton. Imagenet classification with deep convolutional neural networks. In *Neural Inform. Process. Syst.*, 2012. 7

[30] C. Li, K. Zhou, and S. Lin. Intrinsic face image decomposition with human face priors. In *European Conf. Comput. Vision*, pages 218–233. Springer, 2014. 1, 2

[31] I. Masi, A. Tran, T. Hassner, J. T. Leksut, and G. Medioni. Do We Really Need to Collect Millions of Faces for Effective Face Recognition? In *European Conf. Comput. Vision*, 2016. Available www.openu.ac.il/home/hassner/projects/augmented_faces. 5

[32] Y. Nirkin, I. Masi, A. T. Tran, T. Hassner, and G. Medioni. On face segmentation, face swapping, and face perception. In *Int. Conf. on Automatic Face and Gesture Recognition*, 2018. Available www.openu.ac.il/home/hassner/projects/faceswap/. 4, 6

[33] A. Odena, V. Dumoulin, and C. Olah. Deconvolution and checkerboard artifacts. *Distill*, 2016. 4

[34] R. Or-El, G. Rosman, A. Wetzler, R. Kimmel, and A. M. Bruckstein. RGBD-fusion: Real-time high precision depth recovery. In *Proc. Conf. Comput. Vision Pattern Recognition*, pages 5407–5416, 2015. 4

[35] O. M. Parkhi, A. Vedaldi, and A. Zisserman. Deep face recognition. In *Proc. British Mach. Vision Conf.*, 2015. 1, 4, 5

[36] P. Paysan, R. Knothe, B. Amberg, S. Romhani, and T. Vetter. A 3D face model for pose and illumination invariant face recognition. In *Int. Conf. on Advanced Video and Signal based Surveillance*, 2009. 1, 2, 3

[37] P. Prez, M. Gangnet, and A. Blake. Poisson image editing. *Proc. ACM SIGGRAPH Conf. Comput. Graphics*, 22(3):313–318, 2003. 5, 6

[38] E. Richardson, M. Sela, and R. Kimmel. 3d face reconstruction by learning from synthetic data. In *Int. Conf. on 3D Vision*, 2016. 2

[39] E. Richardson, M. Sela, R. Or-El, and R. Kimmel. Learning detailed face reconstruction from a single image. In *Proc. Conf. Comput. Vision Pattern Recognition*, July 2017. 1, 2

[40] S. Romdhani and T. Vetter. Efficient, robust and accurate fitting of a 3D morphable model. In *Proc. Int. Conf. Comput. Vision*, 2003. 2

[41] S. Romdhani and T. Vetter. Estimating 3D shape and texture using pixel intensity, edges, specular highlights, texture constraints and a prior. In *Proc. Conf. Comput. Vision Pattern Recognition*, volume 2, pages 986–993, 2005. 7, 8, 14

[42] J. Roth, Y. Tong, and X. Liu. Unconstrained 3D face reconstruction. In *Proc. Conf. Comput. Vision Pattern Recognition*, June 2015. 2

[43] J. Roth, Y. Tong, and X. Liu. Adaptive 3D face reconstruction from unconstrained photo collections. In *Proc. Conf. Comput. Vision Pattern Recognition*, June 2016. 2

[44] M. Sela, E. Richardson, and R. Kimmel. Unrestricted facial geometry reconstruction using image-to-image translation. In *Proc. Int. Conf. Comput. Vision*, 2017. 1, 2, 3, 6, 7, 8, 14

[45] S. Sengupta, A. Kanazawa, C. D. Castillo, and D. Jacobs. SfSNet: Learning shape, reflectance and illuminance of faces in the wild. *arXiv preprint arXiv:1712.01261*, 2017. 2

[46] M. Simon, E. Rodner, and J. Denzler. Imagenet pre-trained models with batch normalization. *arXiv preprint arXiv:1612.01452*, 2016. 7

[47] S. Suwajanakorn, I. Kemelmacher-Shlizerman, and S. M. Seitz. Total moving face reconstruction. In *European Conf. Comput. Vision*, pages 796–812. Springer, 2014. 2

[48] Y. Taigman, M. Yang, M. Ranzato, and L. Wolf. Deepface: Closing the gap to human-level performance in face verification. In *Proc. Conf. Comput. Vision Pattern Recognition*. IEEE, 2014. 1

[49] H. Tang, Y. Hu, Y. Fu, M. Hasegawa-Johnson, and T. S. Huang. Real-time conversion from a single 2d face image to a 3D text-driven emotive audio-visual avatar. In *Int. Conf. on Multimedia and Expo*, pages 1205–1208. IEEE, 2008. 2

[50] A. Tewari, M. Zollhfer, P. Garrido, H. K. Florian Bernard, P. Prez, and C. Theobalt. Self-supervised multi-level face model learning for monocular reconstruction at over 250 Hz. *arXiv preprint arXiv:1712.02859*, 2017. 2

[51] A. Tewari, M. Zollhofer, H. Kim, P. Garrido, F. Bernard, P. Perez, and C. Theobalt. MoFA: Model-based deep convolutional face autoencoder for unsupervised monocular reconstruction. In *Proc. Int. Conf. Comput. Vision*, Oct 2017. 1, 2

[52] A. Tran, T. Hassner, I. Masi, and G. Medioni. Regressing robust and discriminative 3D morphable models with a very deep neural network. In *Proc. Conf. Comput. Vision Pattern Recognition*, 2017. Available: `http://www.openu.ac.il/home/hassner/projects/CNN3DMM/`. 1, 2, 3, 6, 7, 8, 13, 14

[53] G. Turk and M. Levoy. Zippered polygon meshes from range images. In *Proc. Conf. on Comput. graphics and interactive techniques*, pages 311–318. ACM, 1994. 6

[54] T. Vetter and V. Blanz. Estimating coloured 3d face models from single images: An example based approach. In *European Conf. Comput. Vision*, pages 499–513. Springer, 1998. 1, 2

[55] F. Yang, J. Wang, E. Shechtman, L. Bourdev, and D. Metaxas. Expression flow for 3D-aware face component transfer. *ACM Trans. on Graphics*, 30(4):60, 2011. 2

[56] R. A. Yeh, C. Chen, T. Y. Lim, A. G. Schwing, M. Hasegawa-Johnson, and M. N. Do. Semantic image inpainting with deep generative models. In *Proc. Conf. Comput. Vision Pattern Recognition*, pages 5485–5493, 2017. 5

[57] D. Yi, Z. Lei, S. Liao, and S. Z. Li. Learning face representation from scratch. *arXiv preprint arXiv:1411.7923*, 2014. Available: `http://www.cbsr.ia.ac.cn/english/CASIA-WebFace-Database.html`. 5, 7

[58] Y. Yu, K. Zhou, D. Xu, X. Shi, H. Bao, B. Guo, and H.-Y. Shum. Mesh editing with Poisson-based gradient field manipulation. *ACM Trans. on Graphics*, 23(3):644–651, 2004. 6

[59] X. Zhu, Z. Lei, X. Liu, H. Shi, and S. Z. Li. Face alignment across large poses: A 3D solution. In *Proc. Conf. Comput. Vision Pattern Recognition*, 2016. 1, 2, 3, 7, 8, 14

# A. Supplementary material

## A.1. Overview of our approach

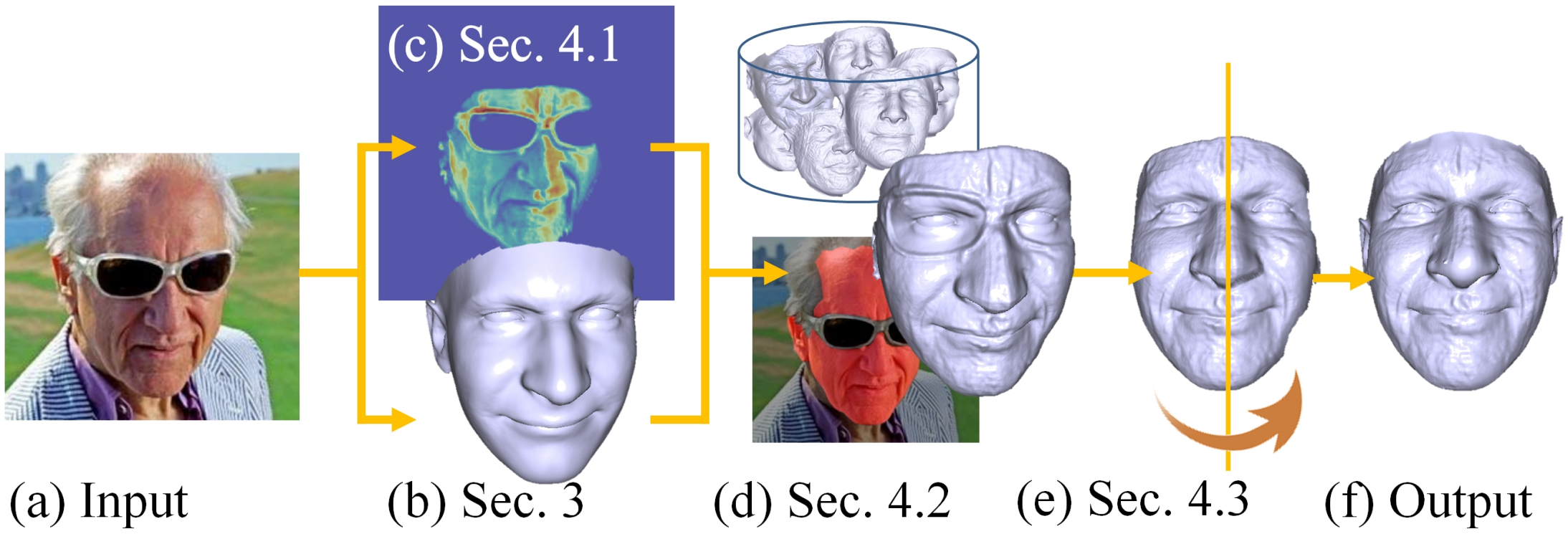

**Figure 9: Method overview (Fig. 2, provided here in higher resolution and with a detailed caption).** (a) Given an input photo (b) we begin by estimating a foundation shape, its expression, and viewpoint. (c) We compute a bump map capturing mid level facial details (visualized here as a heat map) and apply it to the foundation shape. (d) Facial regions are segmented from the background and occlusions (visualized in red over the input image). We use example reference bump maps to complete missing bump map details in occluded facial regions (here, eyes). Finally, (e), we transfer details between symmetric regions of the face to complete self-occluded details. (f) Our final result.

## A.2. Bump map estimation network (addendum to Sec. 4.1)

**Network structure.** Details of our network architecture are provided in Table 2. Our network is similar to the deep encoder-decoder framework with skip connections of [24]. We modify it by adding padding of the input image of 7 pixels. Resolution is decreased using only striding (i.e., *without* pooling layers). The basic block of the network consists of three convolutional layers with a filter size of $3\times3$ and biases, each layer being followed by ReLu activations. The first convolutional layer of each block downsamples by a factor of 2 in the encoder and upsamples by 2 in the decoder. The convolution filters are initialized with the method of [13]; biases are initialized with a constant value of 0.1. Finally, the last decoding block has an additional convolutional layer that maps back to an output of dimension $512\times512\times1$.

**Training process.** The following hyper-parameters were used when training the bump map predicting network. We used Stochastic Gradient Descent (SGD) with batch size set to 4. Learning rate was initialized to $10^{-4}$ and decreased by a factor of 10 once every 50 epochs. The final learning rate was $10^{-6}$. Network training converged after 150 epochs.

| Block | Skip Conn. | Conv. Param. | Output size |
|---|---|---|---|
| | | Encoder ↓ | |
| block1 | | conv3-3, padding = 7<br>[ conv3-16 ] × 2 | 512×512×16 |
| block2 | | conv3-16, stride = 2<br>[ conv3-32 ] × 2 | 256×256×32 |
| block3 | | conv3-32, stride = 2<br>[ conv3-64 ] × 2 | 128×128×64 |
| block4 | | conv3-64, stride = 2<br>[ conv3-128 ] × 2 | 64×64×128 |
| block5 | | conv3-128, stride = 2<br>[ conv3-256 ] × 2 | 32×32×256 |
| block6 | | conv3-256, stride = 2<br>[ conv3-512 ] × 2 | 16×16×512 |
| block7 | | conv3-512, stride = 2<br>[ conv3-1024 ] × 2 | 8×8×1024 |
| | | Decoder ↑ | |
| upblock6 | | upsample+conv3-512 | 16×16×512 |
| | block6 | [ conv3-512 ] × 2 | |
| upblock5 | | upsample+conv3-256 | 32×32×256 |
| | block5 | [ conv3-256 ] × 2 | |
| upblock4 | | upsample+conv3-128 | 64×64×128 |
| | block4 | [ conv3-128 ] × 2 | |
| upblock3 | | upsample+conv3-64 | 128×128×64 |
| | block3 | [ conv3-64 ] × 2 | |
| upblock2 | | upsample+conv3-32 | 256×256×32 |
| | block2 | [ conv3-32 ] × 2 | |
| upblock1 | | upsample+conv3-16 | 512×512×16 |
| | block1 | [ conv3-16 ] × 2 | |
| lastconv | | conv3-1 | 512×512×1 |

**Table 2: Bump map estimation network architecture.**

## A.3. Qualitative ablation study

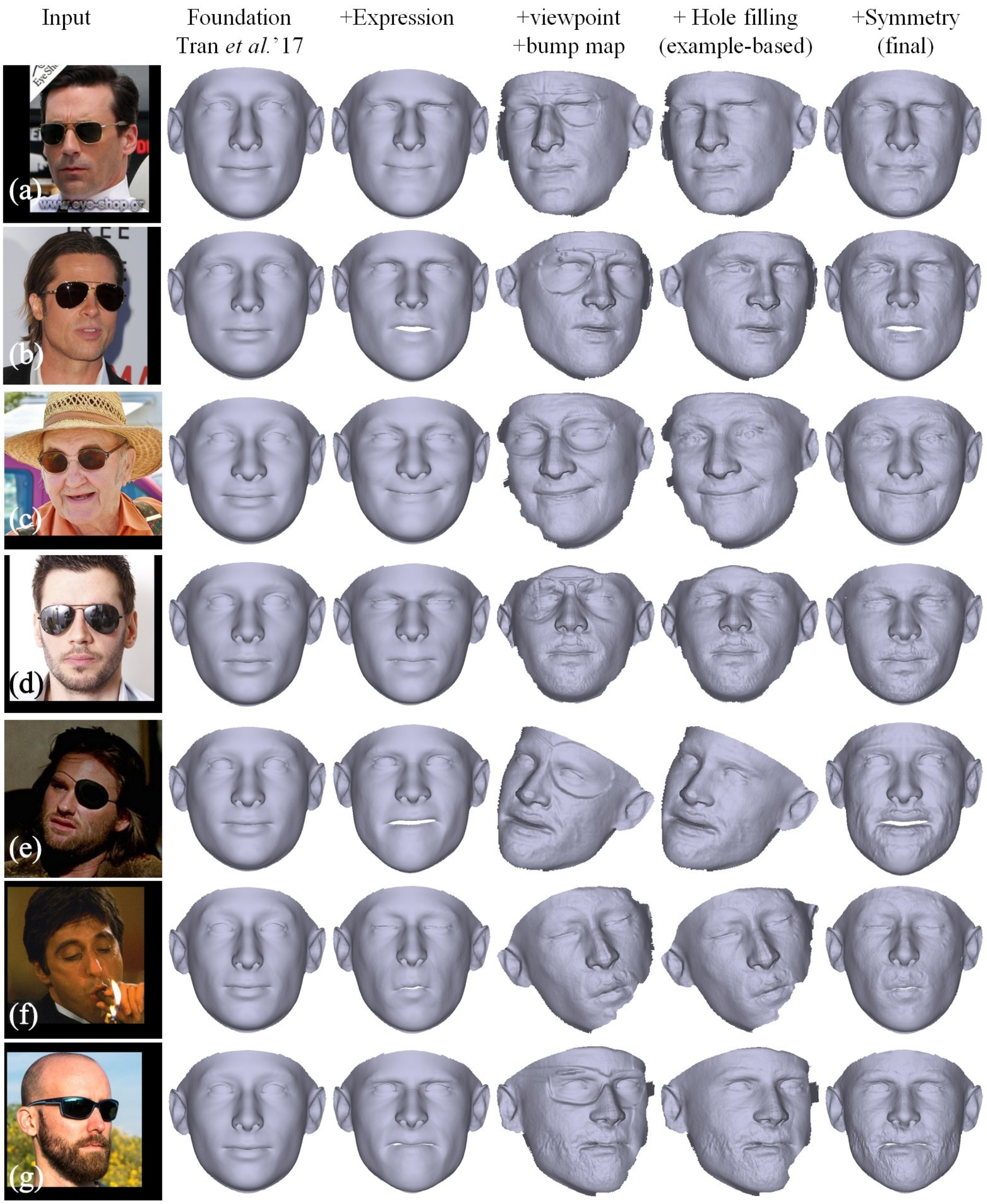

**Figure 10: Qualitative ablation results.** Intermediate 3D reconstruction results of our method. From left to right: The input image; the shape estimated as our foundation using [52] (Sec. 3); with the addition of expression (Sec. 3); aligned to the input viewpoint and layered with an estimated bump map (Sec. 4.1); hole filled using examples (Sec. 4.2); and our final result obtained using soft symmetry to complete self occluded details and rendered to frontal view (Sec. 4.3).

### A.4. Qualitative Comparison

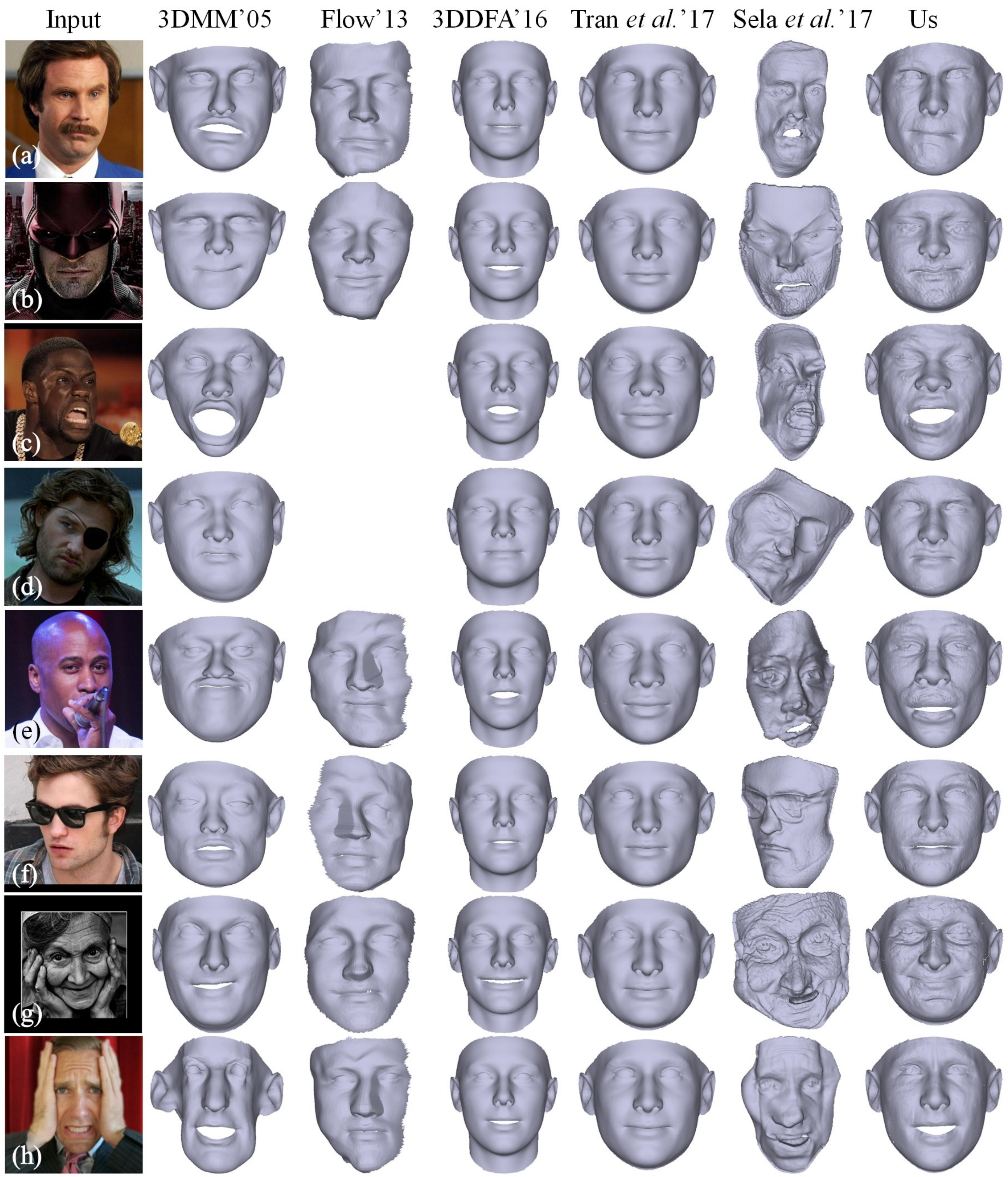

**Figure 11: Qualitative results (addendum to Fig. 8).** Baseline methods from left to right: 3DMM fitting of [41], flow based approach [15] (failed on (c) and (d)), 3DDFA [59], Tran *et al.* [52] (our foundations), and the system of Sela *et al.* [44].